\documentclass{article} 
\usepackage{arxive,times}

\usepackage{amsmath,amsfonts,bm}

\def\eqref#1{equation~\ref{#1}}

\def\1{\bm{1}}

\def\vd{{\bm{d}}}

\DeclareMathAlphabet{\mathsfit}{\encodingdefault}{\sfdefault}{m}{sl}
\SetMathAlphabet{\mathsfit}{bold}{\encodingdefault}{\sfdefault}{bx}{n}

\usepackage{hyperref}
\usepackage{url}
\usepackage{graphicx}
\usepackage{float}
\usepackage[frozencache,cachedir=.]{minted}
\usepackage{array}

\title{AttackQA: Development and Adoption of a Dataset for Assisting Cybersecurity Operations using Fine-tuned and Open-Source LLMs}

\author{Varun Badrinath Krishna\\
SambaNova Systems\\
\texttt{\{first.last\}@sambanova.ai} \\
}

\newcommand{\tab}{\hspace{1cm}}
\iclrfinalcopy 
\begin{document}

\maketitle

\begin{abstract}
Retrieval-augmented generation (RAG) on specialized domain datasets has shown improved performance when large language models (LLMs) are fine-tuned for generating responses to user queries. In this study, we develop a cybersecurity question-answering (Q\&A) dataset, called AttackQA, and employ it to build a RAG-based Q\&A system designed for analysts in security operations centers. The dataset comprises 25,335 Q\&A pairs, accompanied by rationales to facilitate fine-tuning and evaluation. 80\% of the dataset was generated with help of a lightweight open-source LLM (LLama 3 8B), which produced over 1100 tokens per second with full 16-bit precision on SambaNova System's SN40L specialized hardware. To ensure dataset quality, we fine-tuned LLama 3 70B to detect and reject low-quality Q\&A pairs. In using the dataset for RAG, we demonstrate that fine-tuning open-source embeddings and LLMs can yield superior accuracy compared to OpenAI's state-of-the-art proprietary embedding and LLM (GPT-4o). Furthermore, we use Llama 3.1 405B as a judge to evaluate answer correctness, enabling the creation of a fully open-source, high-speed RAG and evaluation pipeline with a  benchmark for model accuracy.
\end{abstract}

\thispagestyle{plain}

\section{Introduction}

Security operations centers (SOCs) house information security teams who are responsible for detecting, investigating, and responding to cybersecurity incidents using a variety of tools, technologies, and processes. As of 2024, firms with at least \$1 billion in revenue spend \$14.6 million on SOCs each year~\citep{c:soccosts} and 80\% of SOC budgets are spent on labor~\citep{c:socbudget}. The cost of training a team of 10 SOC analysts to master 7 security tools is estimated at \$3.69 million~\citep{c:soctraining}. According to the \citet{c:socsurvey} survey, the top challenges facing SOCs include a lack of expertise in security, too much time spent in investigating alerts, and a slow response time to advanced threats. To address those challenges and to enable quicker attack prevention and recovery, we propose a question-answering (Q\&A) system leveraging artificial intelligence to help SOC analysts get quick answers to time-sensitive questions about cyberattacks. Our solution leverages entirely open-source large language models (LLMs) that are becoming increasingly powerful and, on domain-specific datasets, can be tuned to exceed the performance of proprietary LLMs that are many times as large.

We used the MITRE ATT\&CK®~\citep{c:mitre} knowledge base of cyberattack techniques, tools, campaigns, detection approaches, and mitigation approaches to generate a Q\&A dataset called AttackQA for use in Q\&A systems or general-purpose chatbots. That knowledge base, grounded in real-world observations and updated biannually, was chosen because the ATT\&CK® framework is widely adopted for cyber threat intelligence across the private sector, government, and the broader cybersecurity product and service community~\citep{c:mitreuse,c:mitre-use1,c:mitre-use2}. It is stored in an esoteric database format called Structured Threat Information Expression (STIX), making it ill-suited for direct use in Q\&A systems. Hence, we extracted the data and processed it in a way that makes it easier for training and inferencing with LLMs.

\begin{figure}[t]
\centering
\includegraphics[width=0.9\columnwidth]{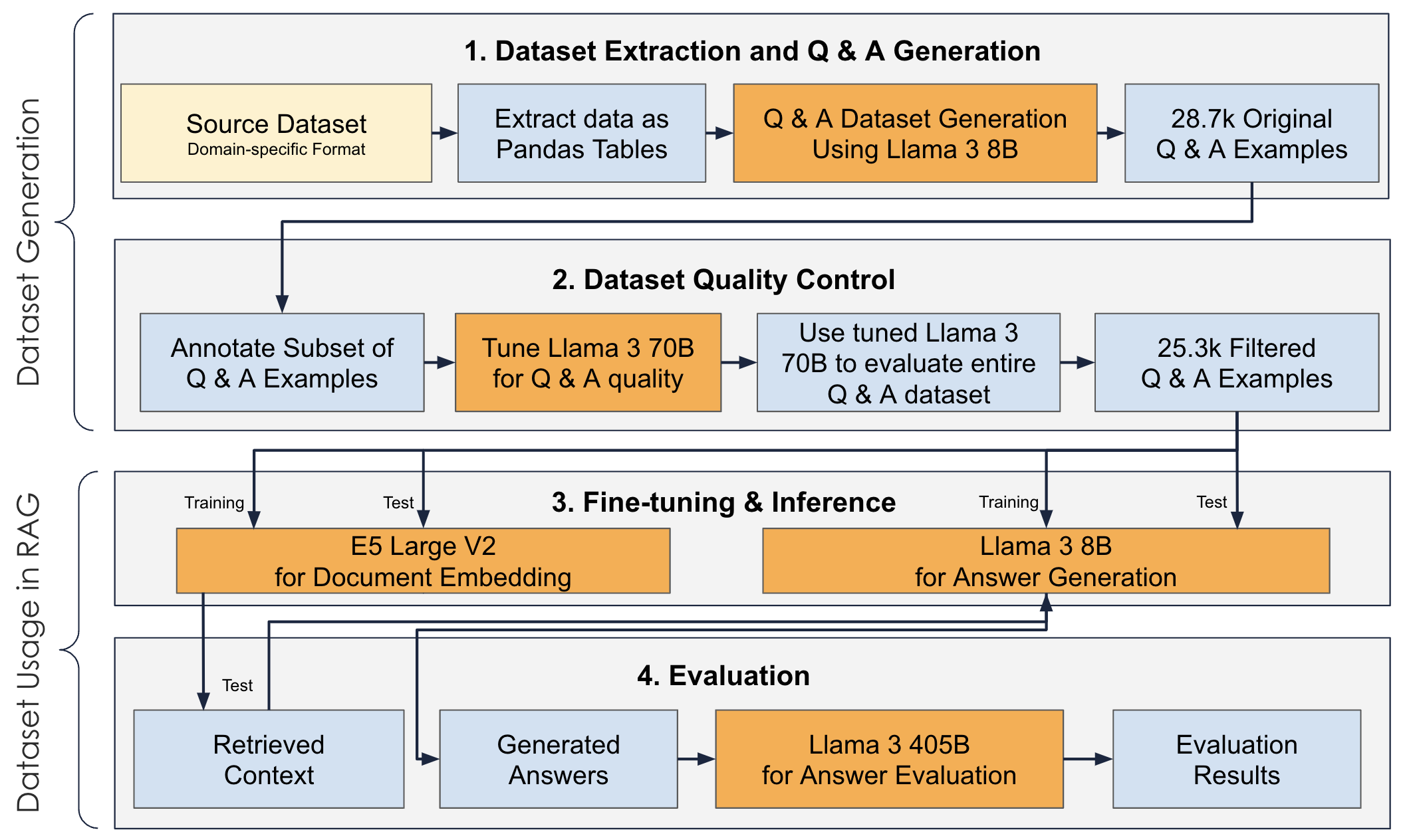} 
\caption{Illustration of dataset generation, quality control, and adoption in RAG}
\label{fig:approach}
\end{figure}

The structure of this paper and our approach are outlined in Fig.~\ref{fig:approach}. The first phase involves the creation of the AttackQA dataset. Initially, we generated 28,686 Q\&A pairs derived from the MITRE knowledge base. Subsequently, we fine-tuned Llama 3 70B to perform quality control (QC) on those Q\&A pairs, retaining 25,335 high-quality examples. In the second phase, AttackQA was used to fine-tune both Microsoft's E5 Large V2 embedding~\citep{c:e5} and Meta's Llama 3 8B LLM~\cite{c:llama3} for retrieval-augmented generation (RAG). The accuracy of the results was assessed using Llama 3 405B, leveraging a G-Eval~\citep{c:geval} correctness metric within the DeepEval Framework.

In summary, our contributions are as follows: 
\begin{itemize} 
    \item We demonstrate the use of a compact, open-source LLM (Llama 3 8B Instruct) to generate a high-quality question-answer dataset from the MITRE ATT\&CK® knowledge base. 
    \item We perform an evaluation that shows that a fine-tuned Llama 3 70B model is better than OpenAI's GPT-4o at identifying questions and answers that are of low quality, so they can be removed from AttackQA as part of an automated dataset quality control process.
    \item We demonstrate that fine-tuning an embedding model significantly enhances context recall in retrieval tasks, outperforming OpenAI's state-of-the-art (SOTA) embedding model, Text-Embedding-3-Large. 
    \item We utilize Llama 3 405B as a judge to evaluate answer correctness. Using its evaluation scores, we found that fine-tuning Llama 3 8B as a generation model in RAG improves correctness, surpassing the performance of OpenAI's GPT-4o, which is many times as large. 
    \item We developed an accurate and low-latency end-to-end RAG pipeline, utilizing fine-tuned open-source embeddings and LLMs to serve as a Q\&A system to support security analysts. 
\end{itemize}

By employing Llama 3 8B at speeds exceeding 1100 tokens/s (at full 16-bit precision), Llama 3 70B at over 550 tokens/s, and Llama 3 405B at 132 tokens/s, we were able to develop a highly responsive end-to-end solution for SOCs. Those model throughputs were achieved using the free SambaNova Cloud platform~\citep{c:sncloud} on specialized hardware~\citep{c:sn40L}.

\section{Related Work}

The use of LLMs for synthetic dataset generation, curation, and evaluation has been surveyed by~\citet{c:synthetic_survey}. Although AttackQA is synthetically generated, it is grounded in the widely reputed MITRE ATT\&CK® knowledge base.

The work of \citet{c:dss} demonstrates that fine-tuning a 770M T5 model, using extracted rationales during the fine-tuning process, can outperform a few-shot prompted 540B PaLM model. Similarly, \citet{c:raft} fine-tuned the generation model within a RAG pipeline, enabling it to predict rationales alongside answers. Moreover, they fine-tuned the model using context that included both relevant and irrelevant (distractor) documents to improve its ability to answer questions. We employ the same set up in our work and show that greater accuracy improvements can be obtained by fine-tuning the embedding in addition to the LLM.

\citet{c:rankrag} fine-tuned large language models to generate answers while simultaneously ranking context based on relevance. \citet{c:atlas} introduced the pretrained LLM Atlas, specifically designed for retrieval-augmented generation, which achieved a 3\% performance improvement over a 540B parameter model, despite having 50 times fewer parameters. Our fine-tuned embeddings perform well on ranking without requiring any unique training approaches and our generation models produce a 9\% improvement on proprietary SOTA models that are much larger.

The fine-tuning of embeddings has been previously shown to enhance performance on tasks involving domain-specific datasets~\citep{c:molbert}. Synthetic dataset generation for the explicit purpose of fine-tuning such embeddings has been explored by~\citet{c:synthetic_msr}.

\section{Dataset Creation for Q\&A}

In this section, we describe our methodology for creating a Q\&A dataset using the MITRE ATT\&CK knowledge base.

\subsection{Summary of the source data}

\begin{table}[t]
\setlength\extrarowheight{4pt}
\caption{Types of Entries in the MITRE ATT\&CK knowledge base}
\begin{center}
\begin{tabular}{{p{0.15\linewidth} | p{0.6\linewidth}| p{0.15\linewidth}}}
\cline{1-3}
\textbf{Dataset} & \textbf{Description}                                                            & \textbf{\# of Entries} \\ \cline{1-3}
Techniques       & Attack techniques                                                               & 637                     \\
Tactics          & Attack tactics. One attack technique may employ multiple tactics                & 14                     \\
Software         & Software tools used by cyber attackers when executing techniques                & 677                  \\
Groups           & Names of well-known attacker groups responsible for well-known attack campaigns & 148               \\
Campaigns        & Attack campaigns that may leverage one or more techniques                       & 28                     \\
Mitigations      & Mitigation strategies used for the attacks                                      & 43               \\
Relationships    & Relationships between techniques, software, groups, campaigns, detection and mitigation approaches  & 17324 \\
\cline{1-3}
\end{tabular}
\end{center}
\label{tbl:mitre_types}
\end{table}

The MITRE ATT\&CK knowledge base encompasses seven categories of information, which are detailed in Table~\ref{tbl:mitre_types} along with their corresponding cardinalities. 

The data for techniques, tactics, software, groups, campaigns, and mitigation approaches include a unique ID, name, description, and URL (an example is provided in Appendix~\ref{sec:source-eg}). From that data, we extracted the descriptions as text documents for use in Q\&A tasks. The relationships table maps a \emph{source type} to a \emph{target type} via a \emph{mapping type}. Source types include `software', `group', `data component', `mitigation strategy', and `campaign', while target types consist of `technique', `software', and `group'. The mapping types include `uses', `detects', `mitigates`, and `attributed-to'. A mapping description was also provided and included in our set of Q\&A documents.

AttackQA was partially generated using manual heuristics, with the remainder produced by LLMs. Each Q\&A pair was derived from a single document, eliminating the need for multi-hop reasoning because comprehensive answers did not require information from multiple documents.

\subsection{Document preprocessing}

Newline characters were removed from within individual documents, ensuring that they appeared only between documents in the final retrieval context presented to the generation model. In all the documents, hyperlinks and special tags were replaced with plain text to ensure that neither the embeddings nor the generation models needed to process special tags that would not be encountered in questions and not be expected in answers.

From each document, one to three triplets of $\{question, thought, answer\}$ were generated, where $thought$ represents the rationale necessary to accurately answer the question. Additional metadata was included in the dataset to enable hybrid retrieval approaches that use a combination of vector search, keyword search, relational querying, etc.

\subsection{Manual Q\&A Generation}

Twenty percent of the Q\&A pairs were generated by humans using heuristics embedded in code, relying solely on the relationships table for their creation. 

The human-generated questions resemble ``What campaigns used attack technique 'T1562.001: Disable or Modify Tools'?" The corresponding answers resemble ``The campaigns that used attack technique 'T1562.001: Disable or Modify Tools' were: 'C0002: Night Dragon', 'C0024: SolarWinds Compromise', 'C0028: 2015 Ukraine Electric Power Attack', 'C0029: Cutting Edge'". Because that answer was not available in any single document in the source dataset, we synthetically generated a document to match the answer. That ensured that the full list of relationships for a given source type, target type, and mapping type were available in a single document for ease of retrieval. To generate the document, it was sufficient to query the relationships table, filtering on the relevant entities (e.g., campaigns, software, techniques, etc.). The questions were generated to ensure comprehensive coverage of source types, target types, and mapping types. Notably, no list of relationships was long enough to cause the answer to exceed 1000 tokens in length.

\begin{table}[t]
\setlength\extrarowheight{4pt}
\caption{AttackQA entry with human-generated question and answer}
\begin{center}
\begin{tabular}{rr}
\hline
\textbf{question}        & \begin{tabular}[c]{@{}r@{}}What campaigns used attack technique 'T1562.001: \\ Disable or Modify Tools'?\end{tabular}                                                                                                                                                                                  \\ \hline
\textbf{thought}         & \begin{tabular}[c]{@{}r@{}}To answer the question, I need to know what campaigns \\ used attack technique 'T1562.001: Disable or Modify Tools'\end{tabular}                                                                                                 \\ \hline
\textbf{answer}          & \begin{tabular}[c]{@{}r@{}}The campaigns that used attack technique \\ 'T1562.001: Disable or Modify Tools' were: \\ 'C0002: Night Dragon', 'C0024: SolarWinds Compromise', \\ 'C0028: 2015 Ukraine Electric Power Attack',\\ 'C0029: Cutting Edge'\end{tabular} \\ \hline
\textbf{document}        & \begin{tabular}[c]{@{}r@{}}The campaigns that used attack technique \\ 'T1562.001: Disable or Modify Tools' were: \\ 'C0002: Night Dragon', 'C0024: SolarWinds Compromise', \\ 'C0028: 2015 Ukraine Electric Power Attack',\\ 'C0029: Cutting Edge'\end{tabular} \\ \hline
\textbf{subject\_id}     & T1562.001                                                                                                                                                                                                                                                   \\ \hline
\textbf{subject\_name}   & Impair Defenses: Disable or Modify Tools                                                                                                                                                                                                                    \\ \hline
\textbf{subject\_type}   & T1562                                                                                                                                                                                                                                                       \\ \hline
\textbf{url}             & https://attack.mitre.org/techniques/T1562/001                                                                                                                                                                                                               \\ \hline
\textbf{source}          & relationships\_campaigns\_for\_technique                                                                                                                                                                                                                    \\ \hline
\textbf{references}      & NaN                                                                                                                                                                                                                                                         \\ \hline
\textbf{human\_question} & True                                                                                                                                                                                                                                                        \\ \hline
\textbf{human\_answer}   & True                                                                                                                                                                                                                                                        \\ \hline
\textbf{field}           & NaN                                                                                                                                                                                                                                                         \\ \hline
\textbf{relation\_id}    & NaN                                                                                                                                                                                                                                                         \\ \hline
\textbf{relation\_name}  & NaN  \\ \hline
\end{tabular}
\end{center}
\label{tbl:attackqa-eg1}
\end{table}

An example entry is presented in Table~\ref{tbl:attackqa-eg1}, for which the question and answer were generated using human-defined heuristics ($human\_question = True$ and $human\_answer = True$). Questions similar to that example would have been difficult for models to answer if we had not constructed documents summarizing the relationships between campaigns and techniques (see the $source$ field). The reason is that, without the summary documents, one document would have needed to be retrieved for each relation and presented to the LLM in the context. If the number of relations exceeded $k$, which was 5 in our case, the LLM would not have had all the information on relations to comprehensively answer the question. An alternative to presenting summaries of relations in a single document is to use an agentic approach, wherein the model has relational capabilities to navigate joins (e.g., text-to-SQL) and it can perform retrieval using those relational queries (e.g., through the use of a tool). 

\subsection{LLM-based Automated Q\&A Generation}
Utilizing Llama 3 8B Instruct~\cite{c:llama3}, we generated Q\&A based on the processed documents. To speed up the process, we leveraged SambaNova's free cloud offering~\citep{c:sncloud}, which runs Llama 3 8B at over 1100 tokens/s with full 16 bit precision~\cite{c:llama31000}. Although we experimented with Llama 3 70B, the results were not significantly different, leading us to continue with Llama 3 8B. In some instances, Llama 3 70B produced overly verbose responses that were difficult to quickly comprehend.

Half of the Q\&A pairs comprised human-generated questions and LLM-generated answers. The questions encompassed those that we anticipated end users would most likely ask and were structured as follows: ``Describe X", ``How can X detect Y?", ``How can X mitigate Y?", and ``How does attack software X use attack technique Y?", where X and Y were obtained from the entities listed in Table~\ref{tbl:mitre_types}. Each of those questions could be answered by a specific document in AttackQA. Therefore, the generation model simply had to summarize that document in response to the question. 

\begin{table}[t]
\setlength\extrarowheight{4pt}
\caption{AttackQA entry with human-generated question and LLM generated answer}
\begin{center}
\begin{tabular}{rr}
\hline
\textbf{question}        & \begin{tabular}[c]{@{}r@{}}How does attack software 'S0467: TajMahal' \\ use attack technique 'T1123: Audio Capture'?\end{tabular}                                                                                                                                                                                                                                                                                                                          \\ \hline
\textbf{thought}         & \begin{tabular}[c]{@{}r@{}}To answer the question, I need to understand how \\ TajMahal, an attack software, utilizes the 'T1123: Audio Capture' \\ technique.\end{tabular}                                                                                                                                                                                                                                        \\ \hline
\textbf{answer}          & \begin{tabular}[c]{@{}r@{}}TajMahal has the ability to capture VoiceIP application \\ audio on an infected host.\end{tabular}                                                                                                                                                                                                                                                                                                                               \\ \hline
\textbf{document}        & \begin{tabular}[c]{@{}r@{}}How attack software 'S0467: TajMahal' uses  \\ attack technique 'T1123: Audio Capture':\\ TajMahal has the ability to capture VoiceIP application \\ audio on an infected host.\end{tabular}                                                                                                                                                                                                \\ \hline
\textbf{subject\_id}     & T1123                                                                                                                                                                                                                                                                                                                                                                                                           \\ \hline
\textbf{subject\_name}   & Audio Capture                                                                                                                                                                                                                                                                                                                                                                                                   \\ \hline
\textbf{subject\_type}   & techniques                                                                                                                                                                                                                                                                                                                                                                                                      \\ \hline
\textbf{url}             & https://attack.mitre.org/techniques/T1123                                                                                                                                                                                                                                                                                                                                                                       \\ \hline
\textbf{source}          & relationships\_uses\_software                                                                                                                                                                                                                                                                                                                                                                                   \\ \hline
\textbf{references}      & \begin{tabular}[c]{@{}r@{}}{[}\{'source': 'T1123/TajMahal: https://attack.mitre.org/techniques/T1123', \\ 'citation':  'TajMahal has the ability to capture \\ VoiceIP application audio on an infected host.'\}, \\ \{'source': 'T1123/TajMahal: https://attack.mitre.org/techniques/T1123', \\ 'citation': 'TajMahal has the ability to capture \\ VoiceIP application audio on an infected host.'\}{]}\end{tabular} \\ \hline
\textbf{human\_question} & True                                                                                                                                                                                                                                                                                                                                                                                                            \\ \hline
\textbf{human\_answer}   & False                                                                                                                                                                                                                                                                                                                                                                                                           \\ \hline
\textbf{field}           & NaN                                                                                                                                                                                                                                                                                                                                                                                                             \\ \hline
\textbf{relation\_id}    & S0467                                                                                                                                                                                                                                                                                                                                                                                                           \\ \hline
\textbf{relation\_name}  & TajMahal                                                                                                                                                                                                                                                                                                                                                                                                        \\ \hline
\end{tabular}
\end{center}
\label{tbl:attackqa-eg2}
\end{table}

An example of an entry, with the question generated using human-defined heuristics ($human\_question = True$) and answer generated using an LLM ($human\_answer = False$), is presented in Table~\ref{tbl:attackqa-eg2}. The $thought$ and $references$ were also generated by the LLM. The references include citations to the document to ensure that the LLM-generated answer is grounded in the document. The questions of this type were generated using heuristics that aimed at mimicking potential user questions, while ensuring that all entities in the knowledge base were covered. In this example, the $subject\_id$ and $subject\_name$ refer to a technique and the $relation\_id$ and $relation\_name$ refer to the related software that the question mentions. In other examples of this type, the subject and relation may refer to other entities mentioned in Table~\ref{tbl:mitre_types}.

The remaining 30\% of Q\&A pairs were generated using Llama 3 8B Instruct, where the question, answer, and rationale were all derived from a given document. Depending on the document length, up to three sets 
of $\{question, thought, answer, references\}$ were generated in a single LLM completion. The precise prompt utilized for that generation is given in Appendix~\ref{sec:prompt_datagen}.

\begin{table}[t]
\setlength\extrarowheight{4pt}
\caption{AttackQA entry with LLM-generated question and answer}
\begin{center}
\begin{tabular}{rr}
\hline
\textbf{question}        & How does Akira initially access victim environments?                                                                                                                                                                                                                                                                                                                                                                                                                                                                                                                                                                                                                                                                                          \\ \hline
\textbf{thought}         & \begin{tabular}[c]{@{}r@{}}To answer the question, I need to understand the initial access \\ mechanism used by Akira as described in the document.\end{tabular}                                                                                                                                                                                                                                                                                                                                                                                                                                                                                                                                                                              \\ \hline
\textbf{answer}          & \begin{tabular}[c]{@{}r@{}}Akira uses compromised credentials to access single-factor \\ external access mechanisms such as VPNs for initial access\end{tabular}                                                                                                                                                                                                                                                                                                                                                                                                                                                                                                                                                                              \\ \hline
\textbf{document}        & \begin{tabular}[c]{@{}r@{}}Description of attack group 'G1024: Akira':\\ Akira is a ransomware variant and ransomware deployment \\ entity active since at least March 2023. \\ Akira uses compromised credentials to access single-factor \\ external access mechanisms such as VPNs for initial access, \\ then various publicly-available tools and techniques \\ for lateral movement. Akira operations are associated with \\ "double extortion" ransomware activity, where data is \\ exfiltrated from victim environments prior to encryption, \\ with threats to publish files if a ransom is not paid. \\ Technical analysis of Akira ransomware indicates \\ multiple overlaps with and similarities to Conti malware.\end{tabular} \\ \hline
\textbf{subject\_id}     & G1024                                                                                                                                                                                                                                                                                                                                                                                                                                                                                                                                                                                                                                                                                                                                         \\ \hline
\textbf{subject\_name}   & Akira                                                                                                                                                                                                                                                                                                                                                                                                                                                                                                                                                                                                                                                                                                                                         \\ \hline
\textbf{subject\_type}   & groups                                                                                                                                                                                                                                                                                                                                                                                                                                                                                                                                                                                                                                                                                                                                        \\ \hline
\textbf{url}             & https://attack.mitre.org/groups/G1024                                                                                                                                                                                                                                                                                                                                                                                                                                                                                                                                                                                                                                                                                                         \\ \hline
\textbf{source}          & groups                                                                                                                                                                                                                                                                                                                                                                                                                                                                                                                                                                                                                                                                                                                                        \\ \hline
\textbf{references}      & \begin{tabular}[c]{@{}r@{}}{[}\{'source': 'groups/G1024/description: \\ https://attack.mitre.org/groups/G1024', \\ 'citation': 'Akira uses compromised credentials to \\ access single-factor external access mechanisms \\ such as VPNs for initial access'\}{]}\end{tabular}                                                                                                                                                                                                                                                                                                                                                                                                                                                                \\ \hline
\textbf{human\_question} & False                                                                                                                                                                                                                                                                                                                                                                                                                                                                                                                                                                                                                                                                                                                                         \\ \hline
\textbf{human\_answer}   & False                                                                                                                                                                                                                                                                                                                                                                                                                                                                                                                                                                                                                                                                                                                                         \\ \hline
\textbf{field}           & description                                                                                                                                                                                                                                                                                                                                                                                                                                                                                                                                                                                                                                                                                                                                   \\ \hline
\textbf{relation\_id}    & NaN                                                                                                                                                                                                                                                                                                                                                                                                                                                                                                                                                                                                                                                                                                                                           \\ \hline
\textbf{relation\_name}  & NaN                                                                                                                                                                                                                                                                                                                                                                                                                                                                                                                                                                                                                                                                                                                                           \\ \hline

\end{tabular}
\end{center}
\label{tbl:attackqa-eg3}
\end{table}

The example entry presented in Table~\ref{tbl:attackqa-eg3}, has both question and answer generated using an LLM ($human\_question = False$ and $human\_answer = False$). The $thought$ and $references$ were also generated by the LLM, similar to the previous example. The $subject$ of the question is an attack group and its description $field$ was used to generated the $document$ (from which the $question$, $thought$, and $answer$ were generated).

\subsection{Ensuring Quality of LLM-generated data}
Through careful prompt engineering and post-processing of special characters (e.g., the `$\backslash$' character in file paths), we achieved a 99\% success rate in ensuring that Llama 3 8B produced valid JSON in the required format. That was despite the present lack of a JSON mode in the SambaNova API.

To ensure the quality of the LLM-generated questions and answers, we employed three strategies:
\begin{enumerate}
    \item We mandated that the LLM generate a citation for each question-answer-rationale entry that included verbatim text from the document supporting the answer. That requirement ensured that the entry was grounded in the source material.
    \item All instances of duplicated questions were removed from the dataset, allowing the remaining questions to act as a unique index. The reason is that the presence of duplicate questions implies that the same inquiry could pertain to two or more different documents, indicating that the question lacked specificity to any particular document. 
    \item We leveraged an independent grader LLM to grade each entry in the dataset on the quality of the question and answer. We refer to that LLM as the quality control (QC) LLM and the remainder of this subsection will describe that automated curation approach.
\end{enumerate}

\subsubsection{Criteria for assessing Q\&A quality}
We employed the G-Eval metric~\citealp{c:geval} in the DeepEval framework to facilitate automatic curation using the quality control (QC) LLM. The G-Eval metric is a custom metric that scores each Q\&A pair by assessing it in conjunction with the retrieved context. 
In addition to generating a score, the prompts formulated by DeepEval require the LLM to provide a rationale for the assigned score. The scoring is based on specific evaluation steps that establish the scoring criteria. We developed distinct metrics for assessing both question quality and answer quality.

For the assessment of question quality, we evaluated whether the questions were ambiguous, failed to reference specific topics in the context, referred to topics not present in the context, or were so broad that they could be answered with information outside of the provided context.

For the assessment of answer quality, we examined whether the answers were irrelevant to the question, did not reference pertinent content from the context, lacked comprehensiveness, were vague, or included information not contained within the context.

In DeepEval, the model is prompted to generate a score ranging from 0 to 10. However, that score is subsequently divided by 10 during post-processing to yield a normalized range from 0 to 1.

\subsubsection{Fine-tuning the QC LLM}

\begin{figure}[h]
\centering
\includegraphics[width=0.5\columnwidth]{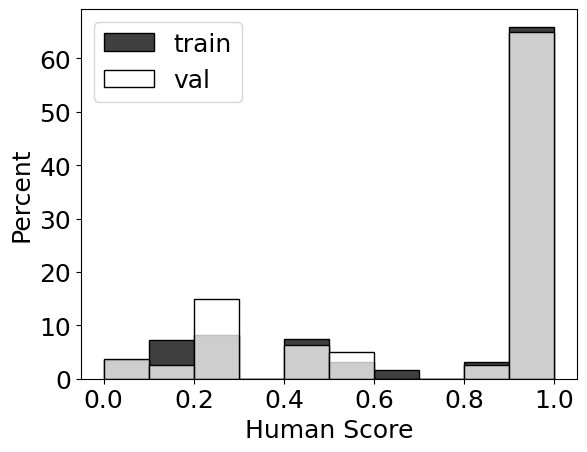} 
\caption{Distribution of scores in the manually-annotated QC dataset}
\label{fig:judging-trainval}
\end{figure}

We manually annotated 400 Q\&A pairs, assigning scores along with reasons to justify the scores. We kept 80 of those pairs as a hold-out validation set to evaluate models ensuring that the score distributions were preserved (see Fig.~\ref{fig:judging-trainval}). A high score indicates that the Q\&A pair is of high quality and worth retaining, whereas pairs with low scores were filtered out. 

According to our annotations, a score of 1 denoted a perfect example, while scores of 0.8 and 0.9 were deemed acceptable, but not flawless. Any score below 0.7 indicated significant quality issues and the example was marked for removal from the dataset. We classified examples from the dataset that are deemed worth retaining (i.e., $score > 0.7$) as positives, and those that are better suited for removal (i.e., $score \le 0.7$) as negatives. Under this classification, $31\%$ of the training set and $32.5\%$ of the validation set were identified as negatives.

At the time of writing, OpenAI's GPT-4o is recognized as a SOTA proprietary model. However, its ability to judge data quality was found to be inadequate. Despite the validation set containing 26 negatives, GPT-4o identified only 2 as negatives, while Llama 3 70B did not predict any negatives. That resulted in $(precision, recall)$ values of $(79\%, 98\%)$ for GPT-4o and $(71\%, 100\%)$ for Llama 3 70B. Neither model demonstrated proficiency in predicting negatives, and the high recall values were largely attributable to the trivial strategy of predicting all examples as positives.

\begin{figure}[h]
\centering
\includegraphics[width=0.5\columnwidth]{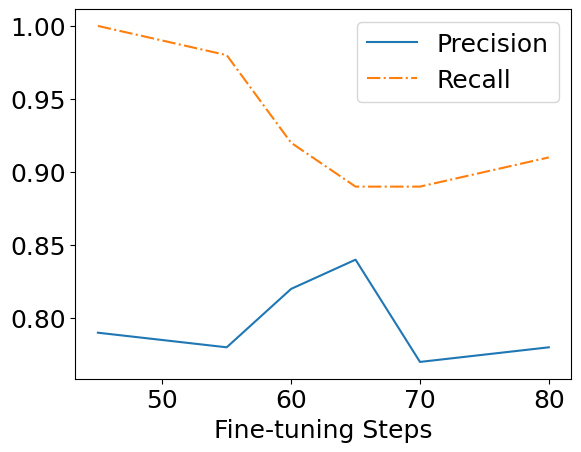} 
\caption{Precision \& recall of QC LLM on annotated validation set}
\label{fig:judging-pr}
\end{figure}

To ensure that the LLM-based QC process would yield a filtered dataset with a high precision (in which case most predicted positives turn out to be true positives), we opted to fine tune Llama 3 70B on the annotated training set. Full parameter tuning was conducted on SambaNova's SambaStudio enterprise platform using the AdamW~\citep{c:adamw} optimizer with a fixed learning rate of $10^{(-5)}$ and a weight decay of $0.1$. As previously mentioned, the base model achieved a perfect recall due to its failure to predict any negatives. However, as we fine-tuned the model over additional steps, we observed a decrease in recall accompanied by an increase in precision, as illustrated in Fig.~\ref{fig:judging-pr}. The final QC assessment was performed using a checkpoint of Llama 3 70B that was fine-tuned for 65 steps, resulting in a validation precision of $84.2\%$ and a recall of $89\%$. By prioritizing precision over recall, we chose a model that was better at detecting negatives (bad examples) and minimizing their presence in the final dataset. The consequent reduction in recall led to the exclusion of some positive examples (along with the negatives) from the final dataset, a trade-off that we preferred because it ensured quality over quantity. 

\subsection{Dataset Summary}
After performing QC, we present AttackQA's summary statistics in Table~\ref{tbl:dataset_stats}. The breakdown of the 25,335 Q\&A pairs by human vs LLM generation is also specified. The token lengths of the documents were measured by the cl100k\_base tokenizer with Tiktoken~\citep{c:tiktoken}.

\begin{table}[h]
\caption{Dataset Summary}
\setlength\extrarowheight{4pt}
\begin{center}
\begin{tabular}{{p{0.4\linewidth} | p{0.4\linewidth}}}
\multicolumn{2}{c}{\textbf{Dataset Summary}}  \\ \cline{1-2}  
Total Number of Q\&A Pairs                   & 25,335       \\
Human Question, Human Answer      & 20\%         \\
Human Question, LLM Answer        & 50\%         \\
LLM Question, LLM Answer          & 30\%        \\ \cline{1-2}  
Number of Unique Documents               & 17,760       \\
Max Document Length               & 3,103 tokens \\
Min Document Length               & 15 tokens    \\
Mean Document Length              & 75 tokens    \\
\cline{1-2}
\end{tabular}
\end{center}
\label{tbl:dataset_stats}
\end{table}

In using the dataset, some of the documents may need to be chunked for use with models with small context windows. Note that only 104 out of 17,760 (0.6\%) of documents have greater than $500$ tokens in length and the rest could be used directly with a model of $4096$ context length. In the analyses presented in the following sections, we did not chunk any of the documents the open-source LLMs we used had context windows of length $8192$ tokens.

\section{Model Fine-tuning for RAG}

\begin{figure}[h]
\centering
\includegraphics[width=0.7\columnwidth]{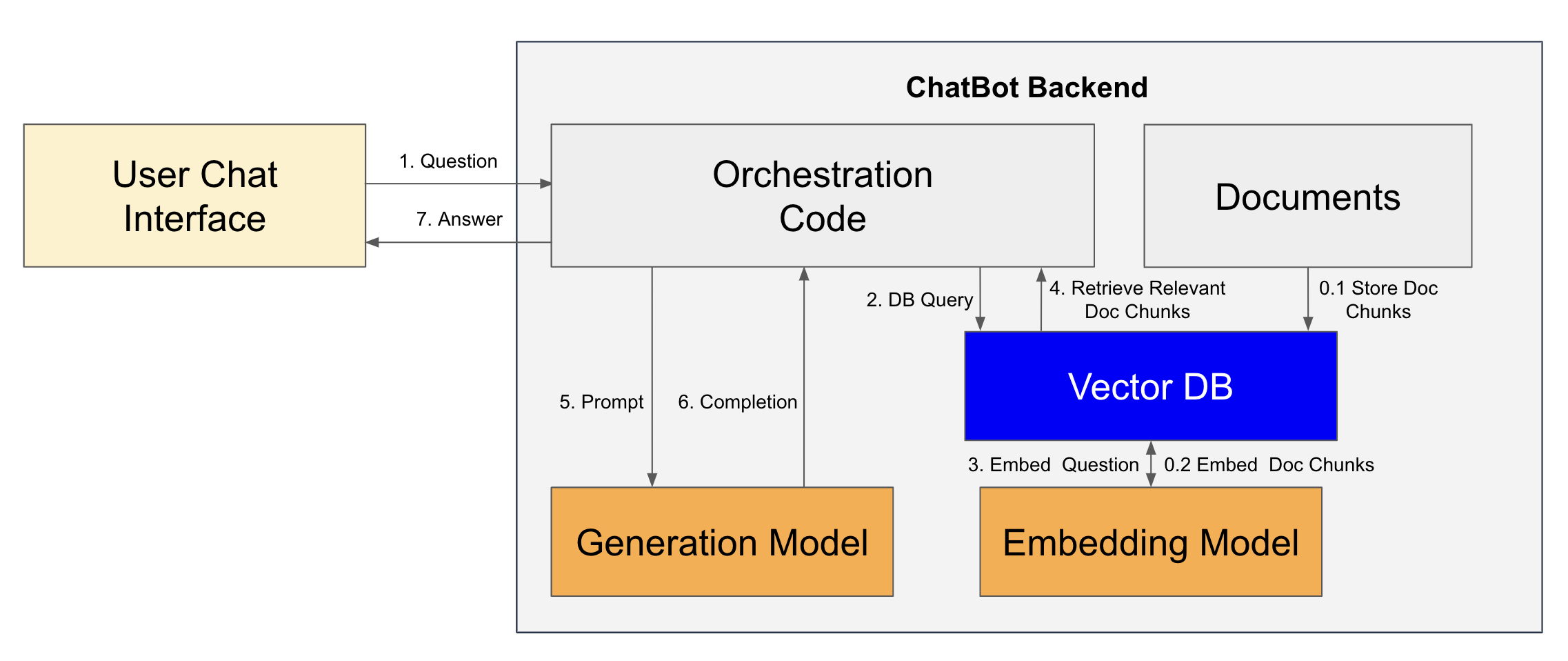} 
\caption{A basic retrieval augmented generation (RAG) framework}
\label{fig:rag}
\end{figure}

A basic RAG framework is illustrated in Fig.~\ref{fig:rag}. In this section, we used AttackQA to fine-tune the LLMs and embeddings using SambaStudio to improve answer accuracy in that framework.

Prior to any user interaction, documents are embedded by the embedding model and stored in a vector database. When a user asks a question, the question is embedded by the same embedding model and $k$ documents relevant to the question are retrieved from the vector database. In our analysis, we retrieve the $k$ documents based on the similarity between their embeddings and the question's embeddings. The $k$ documents are then presented to the generation model in a prompt that asks the model to answer the user's questions using information from the documents. The answer is then returned to the user.

Note that there a more complex implementations of the framework involving multiple generation models, re-ranking models, and multiple types of data stores. Such implementations are beyond on the scope of this analysis, which seeks to measure the contributions of fine-tuning individual models on the overall accuracy.

\subsection{Training and Evaluation Split}

We split the $25,335$ Q\&A pairs into a training set ($90\%$) and an evaluation set ($10\%$) using uniform random sampling. Similar to~\citet{c:raft}, we ensured that all documents were represented in the training set so that the trained models would be familiar with the knowledge base from which questions would be asked. However, the questions in the evaluation set were not present in the training set. That resembles a live production usage setting, in which the end user wants to ask questions of a dataset, and the chatbot is familiar with the source documents but may not have previously seen the questions.

When fine-tuning the models, we used 10\% of the training set for validation to ensure that we would eventually evaluate on a checkpoint that was not over-fitting on the training set.

\subsection{Embedding Model}

We performed full parameter fine-tuning on Microsoft's E5 Large V2 embedding model~\citep{c:e5}, which has 335M parameters and encodes up to 512 tokens into an embedding of length 1024. The training dataset comprised of a list of questions from the training set. For each question, a list of positive documents (containing the answer) and negative documents (not containing the answer) were provided. By construction of the dataset, only one positive document existed for each question (since the question was generated from that document). The dataset was uploaded to SambaStudio and the job was run through the user interface.

Having negative documents helps the model learn to distinguish between relevant and irrelevant documents for a given question using contrastive learning~\citep{c:contrastive}.
The negative documents were randomly sampled from a set that excluded documents whose entities were related to the entity associated with the question. That ensured that the answers could not accidentally be obtained from the negative documents, leading to poor contrastive learning. Related entities can be identified in the MITRE dataset based on their IDs (e.g., T1562.001 and T1562.002 are related techniques and should not be included in negative documents for any question relating to T1562.xxx).

\subsection{Generation Model}
The generation model was fine-tuned on SambaStudio using the same questions that were used to train the embedding model, but the dataset preparation for training was different. For each question, a set of $k$ documents were retrieved using Microsoft's E5 Large V2 embedding. We denote the retrieved set of documents by $\displaystyle \vd(k) = \{d_1,..., d_{k}\}$. 

Let $d^*$ denote the golden document, which contained the answer to the question. For all the 22,802 Q\&A pairs in the training set, we used post-processing to ensure that their corresponding $d^*\in \vd$. Specifically, when $\vd$ did not contain $d^*$ from the retrieval, we used code to replace $d_{k}$ with $d^*$. The remaining $k-1$ documents, $d_i \ne d^*$, were distractor documents that the LLM would need to learn to ignore because the model would be presented with all $k$ documents even in production. We shuffled the $k$ documents in $\vd$ to ensure that the model did not learn to pick $d^*$ based on its retrieval rank in $\vd$, and focused on the contents of the documents instead of the ordering. 

Each prompt comprised of an instruction with a one-shot example, and the retrieved list of documents, $\vd(k)$, with $k=5$. The completions included a thought, answer, and references. The thought was included to ensure that the model's answers were well-reasoned and the references ensured that the right document in the $\vd$ was being used in answering the questions. An example of a prompt-completion pair is given in Appendix~\ref{sec:prompt_genrag}.

We augmented the training set with 3,323 additional examples (amounting to one-eighth of the total training set) to train the model not to hallucinate. In doing so, we re-used questions in the training set for which $d^*\notin \vd$ and had the completions modified with answers that read ``I am sorry, I do not have the answer to the question." and an empty references list.

\section{Model Evaluation}

In this section we present the approach to and results of our model evaluations. All results are presented on the hold-out evaluation set comprising 2,533 examples.

\subsection{Retrieval Model}
The retrieval component of the pipeline refers to steps 2-4 in Fig~\ref{fig:rag}. We evaluated that component using the context recall metric, which captures whether or not the retrieved context contains the golden document. 

\begin{table}[]
\setlength\extrarowheight{4pt}
\caption{Context recall in top $k$ documents for retrieval}
\begin{center}
\begin{tabular}{{ p{0.1\linewidth} | p{0.25\linewidth}| p{0.25\linewidth}| p{0.25\linewidth}}}
\cline{1-4}
\textbf{Top $k$} & \textbf{Base E5 Large V2} & \textbf{Tuned E5 Large V2} & \textbf{OpenAI TE-3-Large} \\
\cline{1-4}
$k=10$ & 72.8\%           & 93.29\%           & 83.58\%                    \\
$k=5$  & 69.84\%          & 92.18\%           & 80.85\%                    \\
$k=1$  & 57.6\%           & 81.48\%           & 65.1\%        \\
\cline{1-4}
\end{tabular}
\end{center}
\label{tbl:context-recall}
\end{table}

Based on a metric, which in our analysis was the similarity metric, a vector database can be configured to return the top $k$ results, $\vd(k,q_i)$, for a given query $q_i$. We seek a metric to check if $d^*(q_i) \in \vd(k,q_i)$ for all $q_i$ and their associated golden documents in the evaluation set. For an evaluation set of $N$ queries, the context recall (denoted by $R$) is computed as follows:

\begin{equation}
R(k) = \frac{1}{N}\sum_{i=1}^{N}{\1_{d^*(q_i) \in \vd(k,q_i)}}
\end{equation}

The results for $k\in \{1, 5, 10\}$ are summarized in Table~\ref{tbl:context-recall}. In all cases, the fine-tuned E5 Large V2 model significantly outperformed both the base E5 Large V2 model and Open AI's SOTA embedding model, Text Embedding 3 Large. The reason is that the dataset contained a lot of domain-specific jargon relating to cybersecurity that the base embedding models were not able to encode. Furthermore, the tuned embedding returned $d^*$ in top 5 ranks in $\vd$ 92.18\% of the time, indicating that a re-ranker model would not have been necessary to bump $d^*$ from the top 10 to the top 5. Finally, the tuned embedding produced $d^*$ at the top rank 81.48\% of the time, indicating strong ranking ability.

\subsection{Generation Model}
Because the answers generated by the generation models are all free-form text, it was difficult to come up with an objective evaluation of their correctness. Objective metrics like Bleu~\citep{c:bleu} and Rouge~\citep{c:rouge} perform N-gram comparisons between expected and actual answers and may not recognize when the two are semantically equivalent if they use different words. For that reason, we used an LLM-as-a-judge to score the answers for correctness.

Once again, we used the G-Eval metric with DeepEval to score answers and provide reasons for the scores. With regard to evaluation criteria, we required that the generated answers be penalized for correctness if they 1) contradicted the true answer, 2) omitted details from the true answer that were relevant to the question, and 3) included irrelevant detail that were not present in the true answer.

\begin{table}[]
\setlength\extrarowheight{4pt}
\caption{Pipeline evaluation of different embedding and generation model configurations}
\resizebox{\columnwidth}{!}{%
\begin{tabular}{rccccccc}
\cline{1-8}
\textbf{}                             & \textbf{\begin{tabular}[c]{@{}c@{}}TE-3-L\\ GPT-4o\end{tabular}} & \textbf{\begin{tabular}[c]{@{}c@{}}Base Emb\\ Base Gen\end{tabular}} & \textbf{\begin{tabular}[c]{@{}c@{}}Base Emb\\ Tuned Gen\end{tabular}} & \textbf{\begin{tabular}[c]{@{}c@{}}Base Emb\\ GPT-4o\end{tabular}} & \textbf{\begin{tabular}[c]{@{}c@{}}Tuned Emb\\ Base Gen\end{tabular}} & \textbf{\begin{tabular}[c]{@{}c@{}}Tuned Emb\\ Tuned Gen\end{tabular}} & \textbf{\begin{tabular}[c]{@{}c@{}}Tuned Emb\\ GPT-4o\end{tabular}} \\
\cline{1-8}
\textbf{\% Context Recall at k=5}        & 80.85                                                                     & 69.84                                                                           & 69.84                                                                            & 69.84                                                                   & \textbf{92.18}                                                                   & \textbf{92.18}                                                                    & \textbf{92.18}                                                           \\
\textbf{\% Answer parsing success} & 99.96                                                                     & 98.18                                                                           & 99.88                                                                            & 99.96                                                                   & 99.53                                                                            & \textbf{100.00}                                                                   & 99.84                                                                    \\
\textbf{\% Correct reference}         & 85.47                                                                     & 74.97                                                                           & 74.97                                                                            & 75.96                                                                   & 91.04                                                                            & 88.91                                                                             & \textbf{92.10}                                                           \\
\textbf{\% Mean Correctness (soft)}   & 81.58                                                                     & 75.10                                                                           & 78.49                                                                            & 82.27                                                                   & 78.60                                                                            & \textbf{88.12}                                                                    & 82.87                                                                    \\
\textbf{\% Mean Correctness (hard)} & 72.38                                                                     & 60.97                                                                           & 70.24                                                                            & 63.99                                                                   & 76.03                                                                            & \textbf{86.07}                                                                    & 79.08                \\
\cline{1-8}
\end{tabular}%
}
\label{tbl:gen-results-all}
\end{table}

We used Llama 3 405B~\citep{c:llama3-405b} for the aforementioned evaluation with DeepEval for its SOTA judging ability~\citep{c:llama405_judging}, speed, and cost (it is provided at 132 tokens/s for free by SambaNova Cloud). Seven combinations of embedding and generation models in the RAG framework were evaluated and the evaluation results are summarized in Table~\ref{tbl:gen-results-all}. ``Base Emb" and ``Tuned Emb" refer to the base and fine-tuned versions of E5 Large V2 embedding model, respectively. ``Base Gen" and ``Tuned Gen" refer to the base and fine-tuned versions of Llama 3 8B generation model, respectively. TE-3-L refers to Open AI's SOTA `Text Embedding 3 Large' model.

The first row of Table~\ref{tbl:gen-results-all} recaps the context recall from Table~\ref{tbl:context-recall} for $k=5$ to show how the other results may be impacted by it. The answer parsing success relates to the generation model's ability to produce JSON-formatted answers with the required fields. That all combinations have at least a 98\% parsing success indicates that the one-shot prompts (given in Appendex~\ref{sec:prompt_genrag}) were adequately engineered to ensure correct completions most of the time. ``\% Correct reference" refers to the number of examples for which the correct reference was produced by the generation model. The references comprise URLs that are included in the retrieved context.

Two correctness scores are provided and both use the same G-Eval metric with Llama 3 405B. In the case of ``mean correctness (soft)", if $d^*(q_i) \notin \vd(k,q_i)$ for any $q_i$ and the generated answer is ``I am sorry, I do not have the answer to the question," then we mark the answer as 100\% accurate. That metric compensates for an inaccurate embedding in the retrieval component, explaining why it makes no difference to the result when we switch the embedding from base to tuned while keeping the same base generation model (either Base Gen or GPT-4o).  

The ``mean correctness (hard)" metric requires that the generated answer match the true answer, regardless of the embedding's retrieval accuracy. No concessions are given for the generation model not admitting to knowing the answer. Therefore, soft correctness scores are higher because some of the answers that were marked as incorrect by hard correctness were forgiven my soft correctness.

The biggest gain on hard correctness, an improvement of 26 percentage points, was achieved when going from a Base Emb/Base Gen combination to a Tuned Emb/Tuned Gen combination. An improvement of 16 percentage points was achieved by swapping out the base embedding with a tuned one, for the same generation model.

Tuning the generation model allows it to correctly answer questions even if the answer is not present in the retrieved context leading to an improvement of 10 percentage points when going from a base generation model to a tuned generation model while keeping the embedding the same.

The first column in Table~\ref{tbl:gen-results-all} refers to a solution using Open AI's SOTA embedding and generation models. On hard correctness, that combination outperforms all other combinations that use the base embedding, but it underperforms those that use the tuned embedding. Therefore, tuning the embedding model is essential to beating proprietary SOTA models using open source SOTA models on our evaluation set.

\subsection{Case Studies}

In this section, we present three evaluation case studies to take a deeper look at the evaluation results. Each case study presents the results for a specific Q\&A pair in the evaluation set. A tabular format is used in which the generated answers came from either GPT-4o, Llama 3 8B (base), or Llama 3 8B (fine-tuned), as specified in the column headers. The row $d^*$ rank in context refers to the ranked position in $\vd$ when the retrieval succeeds (otherwise it reads $d^*\notin \vd$). The hard correctness score and reason were both produced by Llama 3 405B, which we used for judging.

\begin{table}[]
\setlength\extrarowheight{4pt}
\caption{Evaluation Case Study: What is the purpose of KOPILUWAK?}
\resizebox{\columnwidth}{!}{%
\begin{tabular}{rrrrrr}
\hline
\textbf{Question} &
  \multicolumn{5}{l}{What is the purpose of KOPILUWAK?} \\
\textbf{True answer} &
  \multicolumn{5}{l}{KOPILUWAK has been used for victim profiling and C2 since at least 2017.} \\
\textbf{Document} &
  \multicolumn{5}{l}{\begin{tabular}[c]{@{}l@{}}Description of attack software 'S1075: KOPILUWAK': KOPILUWAK is a JavaScript-based  reconnaissance tool that has been used \\ for victim profiling and C2 since at least 2017.\end{tabular}} \\ \hline
\multicolumn{1}{l}{} &
  \textbf{Base Emb, Base Gen} &
  \textbf{Base Emb, Tuned Gen} &
  \textbf{Tuned Emb, Base Gen} &
  \textbf{Tuned Emb, Tuned Gen} &
  \textbf{OpenAI Emb, GPT 4-o} \\ \hline
\textbf{Generated answer} &
  \begin{tabular}[c]{@{}r@{}}I am sorry, I do not \\ have the answer \\ to the question.\end{tabular} &
  \begin{tabular}[c]{@{}r@{}}KOPILUWAK is a \\ command-line utility \\ used to compress and \\ encrypt data collected\\  prior to exfiltration, \\ such as tar.\end{tabular} &
  \begin{tabular}[c]{@{}r@{}}KOPILUWAK is a \\ JavaScript-based \\ reconnaissance tool that \\ has been used for victim \\ profiling and C2 since \\ at least 2017.\end{tabular} &
  \begin{tabular}[c]{@{}r@{}}KOPILUWAK has been \\ used for victim profiling \\ and C2 since at least \\ 2017.\end{tabular} &
  \begin{tabular}[c]{@{}r@{}}I am sorry, I do not \\ have the answer \\ to the question.\end{tabular} \\
\textbf{$d^*$ rank in context} &
  $d^* \notin \vd$ &
  $d^* \notin \vd$ &
  1 &
  1 &
  $d^* \notin \vd$ \\
\textbf{\begin{tabular}[c]{@{}r@{}}Correctness score \\ (hard)\end{tabular}} &
  0.0 &
  0.0 &
  0.8 &
  1.0 &
  0.0 \\
\textbf{Correctness reason} &
  \begin{tabular}[c]{@{}r@{}}The actual output \\ does not provide any \\ relevant information \\ about KOPILUWAK, \\ whereas the expected \\ output provides \\ specific details \\ about its purpose.\end{tabular} &
  \begin{tabular}[c]{@{}r@{}}The actual output provides \\ unrelated information \\ about data compression \\ and encryption, whereas \\ the expected output \\ mentions its use for \\ victim profiling and C2 \\ since 2017,  indicating a \\ contradiction.\end{tabular} &
  \begin{tabular}[c]{@{}r@{}}The actual output provides \\ extra information about the \\ programming language of \\ KOPILUWAK, which is \\ not present in the expected \\ output. However, it does \\ convey the same core \\ information as the expected \\ output regarding the \\ purpose of KOPILUWAK.\end{tabular} &
  \begin{tabular}[c]{@{}r@{}}The actual output \\ matches the expected \\ output exactly, \\ with no omissions \\ or irrelevant details.\end{tabular} &
  \begin{tabular}[c]{@{}r@{}}The actual output does \\ not provide any relevant \\ information about \\ KOPILUWAK, whereas \\ the expected output \\ provides specific details \\ about its purpose.\end{tabular} \\ \hline
\end{tabular}%
}
\label{tbl:case-study}
\end{table}

The case study presented in Table~\ref{tbl:case-study} was among the 30\% of all pairs for which the question and the answer were generated from the document using LLama 3 8B. The configurations with the tuned embedding both returned $d^*$ at the highest rank, which is desirable. The others did not return $d^*$ at all, likely because they were not able to properly embed the domain-specific term ``KOPILUWAK". Consequently, the configurations with the tuned embeddings produced correct answers, whereas the others did not. Llama 3 405B produced a score of 0.8 for the ``Tuned Emb, Base Gen" configuration and its reasoning is clear that the answer includes irrelevant details. The scores for the ``Tuned Emb, Base Gen" and Open AI configurations would have been set to 1 for soft correctness for their admission to not knowing the answer. The ``Base Emb, Tuned Gen" configuration, however, would have received a soft correctness score of 0 for hallucinating.

Table~\ref{tbl:case-study2} presents a case study in which all the embeddings find the $d^*$. Although the OpenAI embedding achieved a higher rank of 1 (all others had a rank of 3), GPT-4o generated a less accurate answer than the tuned generation Llama 3 8B. Like GPT-4o, even the base Llama 3 8B failed to mention that ``testing and debugging" are purposes of the `Office Test' registry key. That is despite the fact that the name of the key implies the purpose and the purpose is explicitly stated in $d^*$. That case study highlights that tuning the generation model with rationales can help improve reasoning.

\begin{table}[H]
\setlength\extrarowheight{4pt}
\caption{Evaluation Case Study: What is the purpose of the `Office Test' Registry key?}
\resizebox{\columnwidth}{!}{%
\begin{tabular}{rrrrrr}
\hline
\textbf{Question} &
  \multicolumn{5}{l}{What is the purpose of the 'Office Test' Registry key?} \\ \hline
\textbf{True answer} &
  \multicolumn{5}{l}{The 'Office Test' Registry key is used by Microsoft to load DLLs for testing and debugging purposes while developing Office applications.} \\ \hline
\textbf{Document} &
  \multicolumn{5}{l}{\begin{tabular}[c]{@{}l@{}}Description of attack technique 'T1137.002: Office Application Startup: Office Test':\textbackslash{}nAdversaries may abuse the Microsoft Office \\ "Office Test" Registry key to obtain persistence on a compromised system. An Office Test Registry location exists that allows a user \\ to specify an arbitrary DLL that will be executed every time an Office application is started. This Registry key is thought to be used \\ by Microsoft to load DLLs for testing and debugging purposes while developing Office applications. This Registry key is not created by \\ default during an Office installation.\textbackslash{}n\textbackslash{}nThere exist user and global Registry keys for the Office Test feature, such as:\textbackslash{}n\textbackslash{}n\\ * HKEY\_CURRENT\_USER\textbackslash{}Software\textbackslash{}Microsoft\textbackslash{}Office test\textbackslash{}Special\textbackslash{}Perf\textbackslash{}n* HKEY\_LOCAL\_MACHINE\textbackslash{}Software\textbackslash{}Microsoft\\ \textbackslash{}Office test\textbackslash{}Special\textbackslash{}Perf\textbackslash{}n\textbackslash{}nAdversaries may add this Registry key and specify a malicious DLL that will be executed whenever an \\ Office application, such as Word or Excel, is started.\end{tabular}} \\ \hline
\textbf{} &
  \textbf{Base Emb, Base Gen} &
  \textbf{Base Emb, Tuned Gen} &
  \textbf{Tuned Emb, Base Gen} &
  \textbf{Tuned Emb, Tuned Gen} &
  \textbf{OpenAI Emb, GPT 4-o} \\ \hline
\textbf{Generated answer} &
  \begin{tabular}[c]{@{}r@{}}Adversaries may abuse \\ the Microsoft Office\\  'Office Test' Registry key \\ to obtain persistence on \\ a compromised system.\end{tabular} &
  \begin{tabular}[c]{@{}r@{}}The 'Office Test' Registry \\ key is used by Microsoft \\ to load DLLs for testing \\ and debugging purposes \\ while developing Office \\ applications.\end{tabular} &
  \begin{tabular}[c]{@{}r@{}}Adversaries may abuse \\ the Microsoft Office \\ 'Office Test' Registry key\\  to obtain persistence on \\ a compromised system.\end{tabular} &
  \begin{tabular}[c]{@{}r@{}}The 'Office Test' Registry \\ key is used by Microsoft to \\ load DLLs for testing and \\ debugging purposes while \\ developing Office \\ applications.\end{tabular} &
  \begin{tabular}[c]{@{}r@{}}The purpose of the 'Office \\ Test' Registry key is to \\ allow a user to specify an \\ arbitrary DLL that will be \\ executed every time an \\ Office application is \\ started.\end{tabular} \\
\textbf{$d^*$ rank in context} &
  3 &
  3 &
  3 &
  3 &
  1 \\
\textbf{\begin{tabular}[c]{@{}r@{}}Correctness score\\ (hard)\end{tabular}} &
  0.0 &
  1.0 &
  0.0 &
  1.0 &
  0.4 \\
\textbf{Correctness reason} &
  \begin{tabular}[c]{@{}r@{}}The actual output \\ contradicts the expected \\ output, stating it's used \\ for persistence by \\ adversaries instead of \\ testing and debugging \\ purposes by Microsoft.\end{tabular} &
  \begin{tabular}[c]{@{}r@{}}The actual output \\ perfectly matches the \\ expected output, \\ providing the same \\ information without \\ any omissions or \\ contradictions.\end{tabular} &
  \begin{tabular}[c]{@{}r@{}}The actual output \\ contradicts the expected \\ output, stating it's used \\ for persistence by \\ adversaries instead of \\ testing and debugging \\ purposes by Microsoft.\end{tabular} &
  \begin{tabular}[c]{@{}r@{}}The actual output \\ perfectly matches the \\ expected output, providing \\ the same information \\ without any omissions \\ or contradictions.\end{tabular} &
  \begin{tabular}[c]{@{}r@{}}The actual output \\ provides a partially correct \\ purpose of the 'Office Test' \\ Registry key but includes \\ arbitrary user specification \\ which is not present in the \\ expected output, and omits \\ the testing and debugging \\ purposes mentioned in the \\ expected output.\end{tabular}
\end{tabular}%
}
\label{tbl:case-study2}
\end{table}

Table~\ref{tbl:case-study3} presents a case study in which the `Base Emb, Tuned Gen' configuration is able to answer a question accurately even in the absence of $d^*$. That shows that fine-tuning the generation model on questions that cover a document, can allow the model to answer unseen questions about that document even when it is not presented in the LLM prompt context.

\begin{table}[H]
\setlength\extrarowheight{4pt}
\caption{Evaluation Case Study: When was MoleNet first observed in use?}
\resizebox{\columnwidth}{!}{%
\begin{tabular}{rrrrrr}
\hline
\textbf{Question} &
  \multicolumn{5}{l}{When was MoleNet first observed in use?} \\ \hline
\textbf{True answer} &
  \multicolumn{5}{l}{MoleNet has been observed in use since at least 2019.} \\ \hline
\textbf{Document} &
  \multicolumn{5}{l}{\begin{tabular}[c]{@{}l@{}}Description of attack software 'S0553: MoleNet':\\MoleNet is a downloader tool with backdoor capabilities that has been observed in use since at least 2019.\end{tabular}} \\ \hline
\multicolumn{1}{l}{} &
  \textbf{Base Emb, Base Gen} &
  \textbf{Base Emb, Tuned Gen} &
  \textbf{Tuned Emb, Base Gen} &
  \textbf{Tuned Emb, Tuned Gen} &
  \textbf{OpenAI Emb, GPT 4-o} \\ \hline
\textbf{Generated answer} &
  \begin{tabular}[c]{@{}r@{}}I am sorry, I do not have\\  the answer to the question.\end{tabular} &
  \begin{tabular}[c]{@{}r@{}}MoleNet was first \\ observed in use in 2019.\end{tabular} &
  \begin{tabular}[c]{@{}r@{}}MoleNet is a downloader \\ tool with backdoor \\ capabilities that has \\ been observed in use \\ since at least 2019.\end{tabular} &
  \begin{tabular}[c]{@{}r@{}}MoleNet has been observed \\ in use since at least 2019.\end{tabular} &
  \begin{tabular}[c]{@{}r@{}}I am sorry, I do not have \\ the answer to the question.\end{tabular} \\
\textbf{$d^*$ rank in context} &
  NaN &
  NaN &
  0.0 &
  0.0 &
  NaN \\
\textbf{Correctness score (hard)} &
  0.0 &
  0.8 &
  0.9 &
  1.0 &
  0.0 \\
\textbf{Correctness reason} &
  \begin{tabular}[c]{@{}r@{}}The actual output does not \\ provide any information \\ about when MoleNet was \\ first observed in use, \\ whereas the expected \\ output states it has been \\ observed since at least \\ 2019.\end{tabular} &
  \begin{tabular}[c]{@{}r@{}}The actual output \\ provides a specific year \\ that matches the \\ expected output, but \\ it implies a specific start \\ date, whereas the \\ expected output \\ leaves room for earlier \\ usage with 'at least'.\end{tabular} &
  \begin{tabular}[c]{@{}r@{}}The actual output \\ provides the correct \\ year MoleNet was first \\ observed in use, but \\ also includes irrelevant \\ details about it being a \\ downloader tool with \\ backdoor capabilities.\end{tabular} &
  \begin{tabular}[c]{@{}r@{}}The actual output matches \\ the expected output exactly, \\ with no omissions or \\ contradictions.\end{tabular} &
  \begin{tabular}[c]{@{}r@{}}The actual output does \\ not provide any \\ information about when \\ MoleNet was first \\ observed in use, whereas \\ the expected output \\ states it has been \\ observed since at least \\ 2019.\end{tabular} \\ \hline
\end{tabular}%
}
\label{tbl:case-study3}
\end{table}

\section{Conclusion}

In this work, we created a Q\&A dataset based off the MITRE ATT\&CK® database of cyberattack techniques, software, campaigns, mitigation approaches, and detection approaches. The dataset, AttackQA, can used to train models and create a chatbot to help security operations center analysts decrease their time to mitigate cyberattacks by giving them fast and accurate answers to questions that they may have about the attacks. We presented an approach to automatically generate data and perform quality control on that data using SOTA open-source LLMs.

We evaluated a RAG pipeline using our dataset and showed that fine-tuning both the generation and embedding models can lead to an increase in hard accuracy of 26 percentage points. Fine-tuning the embedding model alone can lead to an improvement of 16 percentage points. Finally, fine-tuning the generation model alone, as proposed by \citet{c:raft}, leads to an accuracy improvement of 10 percentage points. Open AI's SOTA models produced high accuracy but could be outperformed by tuning openly available embedding models. Even when GPT-4o was combined with our tuned embeddings, it underperformed a fine-tuned Llama 3 8B model, which was many times smaller.

The results establish a benchmark for modeling with AttackQA. The AttackQA dataset and associated benchmarking code are made openly available~\citep{c:attackQALink}.

\subsubsection*{Acknowledgments}
The author would like to thank Amit Kushwaha, Chen Wu, Meenakshi Swaminathan, James Valentine, and Nidhi Hiremath for helping improve the presentation of content in this paper. 

The AttackQA dataset is derived from the MITRE ATT\&CK® knowledge base, which bears the following copyright notice: © 2024 The MITRE Corporation. This work is reproduced and distributed with the permission of The MITRE Corporation.

\bibliography{arxive.bib}
\bibliographystyle{iclr2025_conference}

\newpage
\appendix
\section{Appendix}

\subsection{Examples of tables from source dataset (MITRE)}
\label{sec:source-eg}

In this appendix we present two examples of entries extracted from the MTIRE knowledge base. The first is an example of a software tool used by attackers and is presented in Table~\ref{tbl:software-eg}. There are 677 such entries in the MITRE knowledge base. Entries for techniques, tactics, groups, campaigns, and mitigation approaches also include unique ID, name, description, and URL. 

\begin{table}[H]
\caption{Example software table entry from source data}
\begin{tabular}{{p{0.35\linewidth} | p{0.6\linewidth}}}
\textbf{Field}         & \textbf{Value}                                                                                                                                                                                                                   \\ \hline
ID                     & S0066                                                                                                                                                                                                                               \\
name                   & 3PARA RAT                                                                                                                                                                                                                        \\
description            & {[}3PARA RAT{]}(https://attack.mitre.org/software/S0066 ) is a remote access tool (RAT) programmed in C++ that has been used by {[}Putter Panda{]}(https://attack.mitre.org/groups/G0024 ). (Citation: CrowdStrike Putter Panda) \\
url                    & https://attack.mitre.org/software/S0066                                                                                                                                                                                          \\
contributors           & NaN                                                                                                                                                                                                                              \\
platforms              & Windows                                                                                                                                                                                                                          \\
aliases                &                                                                                                                                                                                                                                  \\
type                   & malware                                                                                                                                                                                                                          \\
relationship citations & (Citation: CrowdStrike Putter Panda),,(Citation: CrowdStrike Putter Panda)                                                                                                                                                      
\end{tabular}
\label{tbl:software-eg}
\end{table}

In creating AttackQA, we preprocess descriptions like ``{[}3PARA RAT{]}(https://attack.mitre.org/software/S0066 ) is a remote access tool (RAT) programmed in C++ that has been used by {[}Putter Panda{]}(https://attack.mitre.org/groups/G0024 ). (Citation: CrowdStrike Putter Panda)" to ``3PARA RAT is a remote access tool (RAT) programmed in C++ that has been used by Putter Panda."

An example of a relationship entry in the MITRE knowledge base is presented in Table~\ref{tbl:relationships-eg}.

\begin{table}[H]
\setlength\extrarowheight{4pt}
\caption{Example relationships table entry from source data}
\begin{tabular}{{p{0.35\linewidth} | p{0.6\linewidth}}}
\textbf{Field}      & \textbf{Value}                                                                                                                                                                                                                                                                                                                                                                                                                                                       \\ \hline
source ID           & M1036                                                                                                                                                                                                                                                                                                                                                                                                                                                                \\
source name         & Account Use Policies                                                                                                                                                                                                                                                               \\
source type         & mitigation                                                                                                                                                                                                                                                                                                                                                                                                                                                           \\
mapping type        & mitigates                                                                                                                                                                                                                                                                                                                                                                                                                                                            \\
target ID           & T1110                                                                                                                                                                                                                                                                                                                                                                                                                                                                \\
target name         & Brute Force                                                                                                 \\
target type         & technique                                                                                                                                                                                                                                                                                                                                                                                                                                                            \\
mapping description & Set account lockout policies after a certain number of failed login attempts to prevent passwords from being guessed. Too strict a policy may create a denial of service condition and render environments un-usable, with all accounts used in the brute force being locked-out. Use conditional access policies to block logins from non-compliant devices or from outside defined organization IP ranges.(Citation: Microsoft Common Conditional Access Policies)
\end{tabular}
\label{tbl:relationships-eg}
\end{table}

\newpage
\subsection{Example Prompts}

In this section, we present the exact prompts that were used for dataset generation and for the RAG application.

\subsubsection{Prompts used for Dataset Generation}
\label{sec:prompt_datagen}

The Python function used to construct the prompts for dataset generation using LLMs is presented below. Valid JSON, containing up to three entries, was generated 99\% of the time even when the API did not support token sampling constraints for JSON outputs. The prompt template is customized for Llama 3 models.

\begin{minted}[
 breaklines,
 breaksymbolleft=
 ]{python}
def get_prompt_for_doc(doc, count="three sets"):
    prompt = """<|begin_of_text|><|start_header_id|>system<|end_header_id|>
    You are a JSON generator who generates machine-readable JSON<|eot_id|><|start_header_id|>user<|end_header_id|>
            Based on the following document, follow the instruction below
            Document:
            %s
            Instruction:
            Generate %s of unique question, thought, answer, and references from the above document in the following JSON format. The answers must avoid words that are not specific (e.g., "many", "several", "few", etc.). The answers must contain specific, verbose, self-contained, grammatically correct sentences that answer the question comprehensively. The answers must strictly contain content from the document and no content from outside the document. There may be multiple references that contain verbatim text from the document to support the answers.             
            JSON format:
            [
                {
                    "question": "<generated question>",            
                    "thought": "<generated thought on what is needed to answer the question. Start with 'To answer the question, I need'>",
                    "answer": "<generated answer>",
                    "references": [
                        "<verbatim text from document that supports the answer>",
                        "<verbatim text from document that supports the answer>"
                    ]
                }
            ]
            The first character of the response must be '[' and the last character must be ']'. No header text should be included.
            <|eot_id|><|start_header_id|>JSON list<|end_header_id|>
            """%(doc, count)
    return prompt
\end{minted}

\newpage
\subsubsection{Prompts for generating answers in RAG}
\label{sec:prompt_genrag}
The RAG prompt contains instructions, a one-shot example to illustrate the required response format, and the entire list of $k=5$ documents from the retrieval model. The prompt template was used for both fine-tuning the model and for inference on the base or fine-tuned models. The tags in the example prompt, which are specific to Llama 3, were removed when performing inference using GPT-4o.

\textbf{Prompt:}
\vspace{4pt}
\hrule
\vspace{4pt}
\textless{}|begin\_of\_text|\textgreater{}\textless{}|start\_header\_id|\textgreater{}system\textless{}|end\_header\_id|\textgreater\\ You are an assistant for generating JSON formatted responses\\ \textless{}|eot\_id|\textgreater{}\textless{}|start\_header\_id|\textgreater{}user\textless{}|end\_header\_id|\textgreater\\ Respond with a JSON dictionary that includes a thought, answer, and references\\ The answer must contain text obtained strictly from the given documents.\\ Avoid any text that is not in the given documents.\\ Answer using concise, self-contained, grammatically complete sentences. \\ The answer must be a string with less than four sentences.\\ Do not mention the documents by number or the context in the answers.\\ Answer the question strictly using the provided documents.\\ If you cannot answer the question using the documents, the answer should be "I am sorry, I do not have the answer to the question."\\ Along with the answer, include a thought that begins with "To answer the question, I need".\\ The references must contain URLs that exactly match the full URLs in the document headers relevant to by the answer.\\ There may be multiple references in the references list.\\ Follow the example below:\\ \\ Document 1:  https://attack.mitre.org/techniques/T1562/001\\ \\ The campaigns that used attack technique 'T1562.001: Disable or Modify Tools' were: 'C0002: Night Dragon', 'C0024: SolarWinds Compromise', 'C0028: 2015 Ukraine Electric Power Attack', 'C0029: Cutting Edge'"\\ 
\\ Document 2: https://attack.mitre.org/techniques/T1562/002\\ \\ The campaigns that used attack technique 'T1562.002: Disable Windows Event Logging' were: 'C0024: SolarWinds Compromise', 'C0025: 2016 Ukraine Electric Power Attack'\\ 
\\ Document 3: https://attack.mitre.org/techniques/T1070/001\\ \\ The campaigns that used attack technique 'T1070.001: Clear Windows Event Logs' were: 'C0014: Operation Wocao'\\ \\ Question: What campaigns used attack technique 'T1562.002: Disable Windows Event Logging'?\\ Response:\\
\{     \\
\tab ``thought": ``To answer the question, I need to know what campaigns used attack technique 'T1562.002: Disable Windows Event Logging'. The answer is contained in the provided document with URL '{https://attack.mitre.org/techniques/T1562/002}'.",\\
    ``answer": ``The campaigns that used attack technique 'T1562.002: Disable Windows Event Logging' were: 'C0024: SolarWinds Compromise', 'C0025: 2016 Ukraine Electric Power Attack'",\\
    ``references": \{[``https://attack.mitre.org/techniques/T1562/002"]\}\\
\}
\\\\

Document 1: https://attack.mitre.org/techniques/T1539 \\
How data component 'Process Access' can be used to detect attack technique 'T1539: Steal Web Session Cookie': \\ Monitor for attempts by programs to inject into or dump browser process memory.\\\\
Document 2: \\ https://attack.mitre.org/techniques/T1539\\ The following 2 data components can be used to detect attack technique 'T1539: Steal Web Session Cookie': File Access, Process Access\\\\
Document 3: https://attack.mitre.org/techniques/T1539\\  The software procedures that use attack technique 'T1539: Steal Web Session Cookie' are: 'S0467: TajMahal', 'S0492: CookieMiner', 'S0531: Grandoreiro', 'S0568: EVILNUM', 'S0631: Chaes', 'S0650: QakBot', 'S0657: BLUELIGHT', 'S0658: XCSSET'\\\\
Document 4:  https://attack.mitre.org/techniques/T1539\\  Tactics used in attack technique 'T1539: Steal Web Session Cookie': Credential Access\\ \\ Document 5:  https://attack.mitre.org/techniques/T1539\\  Description of attack technique 'T1539: Steal Web Session Cookie': An adversary may steal web application or service session cookies and use them to gain access to web applications or Internet services as an authenticated user without needing credentials. Web applications and services often use session cookies as an authentication token after a user has authenticated to a website.  Cookies are often valid for an extended period of time, even if the web application is not actively used. Cookies can be found on disk, in the process memory of the browser, and in network traffic to remote systems. Additionally, other applications on the targets machine might store sensitive authentication cookies in memory (e.g. apps which authenticate to cloud services). Session cookies can be used to bypasses some multi-factor authentication protocols.  There are several examples of malware targeting cookies from web browsers on the local system. Adversaries may also steal cookies by injecting malicious JavaScript content into websites or relying on User Execution by tricking victims into running malicious JavaScript in their browser.  There are also open source frameworks such as `Evilginx2` and `Muraena` that can gather session cookies through a malicious proxy (e.g., Adversary-in-the-Middle) that can be set up by an adversary and used in phishing campaigns.  After an adversary acquires a valid cookie, they can then perform a Web Session Cookie technique to login to the corresponding web application.\\ \\ Question: What is a potential indicator of the 'T1539: Steal Web Session Cookie' attack technique?\\ The response must be formatted as a JSON instance that conforms to the JSON schema above.\\ No text should appear before or after the JSON instance.\\ Response:\\ \textless{}|eot\_id|\textgreater{}\textless{}|start\_header\_id|\textgreater{}machine-readable JSON\textless{}|end\_header\_id|\textgreater{}
\vspace{4pt}
\hrule
\vspace{4pt} 

\textbf{Completion:}
\vspace{4pt}
\hrule
\vspace{4pt}
\{\\
"thought": "To answer the question, I need to understand how to detect the 'T1539: Steal Web Session Cookie' attack technique. The answer is contained in the provided document with URL 'https://attack.mitre.org/techniques/T1539'.",\\ 
"answer": "A potential indicator of the 'T1539: Steal Web Session Cookie' attack technique is monitoring for attempts by programs to inject into or dump browser process memory.",\\ "references": {[}"https://attack.mitre.org/techniques/T1539"{]} \\\}            
\vspace{4pt}
\hrule
\vspace{4pt}
The above completion contains a $thought$, $answer$, and $references$. It is only used for fine-tuning the generation model and is constructed using fields from AttackQA. The $thought$ describes the rationale and is included to help the model learn to find the right document and use it to answer the question.
\end{document}